\documentclass[conference]{IEEEtran}
\IEEEoverridecommandlockouts
\usepackage{cite}
\usepackage{amsmath,amssymb,amsfonts}
\usepackage{algorithmic}
\usepackage{graphicx}
\usepackage[margin=0.75in]{geometry}

\usepackage{hyperref}       % pdf links to e.g. figures and citations
\usepackage{url}            % web links (URL)
\usepackage{float}			% place the object on exact place (figure/table  using [H])
\usepackage{textcomp}
\usepackage[caption=false]{subfig}
\def\BibTeX{{\rm B\kern-.05em{\sc i\kern-.025em b}\kern-.08em
    T\kern-.1667em\lower.7ex\hbox{E}\kern-.125emX}}
\begin{document}

\title{Faster Bounding Box Annotation for Object Detection in Indoor Scenes}

%\author{\IEEEauthorblockN{Bishwo Adhikari}
%\IEEEauthorblockA{\textit{Laboratory of Signal Processing} \\

%\IEEEauthorblockN{Heikki Huttunen}
%\IEEEauthorblockA{\textit{Laboratory of signal Processing} \\
%\textit{Tampere University of Technology}\\
%Tampere, Finland \\
%heikki.huttunen@tut.fi}
%}

\author{\IEEEauthorblockN{$^*$Bishwo Adhikari, $^*$Jukka Peltom\"aki, $^{**}$Jussi Puura and
$^*$Heikki Huttunen}
\IEEEauthorblockA{$^*$Tampere University of Technology, Tampere, Finland\\
$^{**}$Sandvik Mining and Construction Oyj, Tampere, Finland
%Email: \{firstname.surname\}@tut.fi}
}
}

\maketitle

\begin{abstract}
This paper proposes an approach for rapid bounding box annotation for object detection datasets. The procedure consists of two stages: The first step is to annotate a part of the dataset manually, and the second step proposes annotations for the remaining samples using a model trained with the first stage annotations. We experimentally study which first/second stage split minimizes to total workload.
In addition, we introduce a new fully labeled object detection dataset collected from indoor scenes. Compared to other indoor datasets, our collection has more class categories, different backgrounds, lighting conditions, occlusion and high intra-class differences. We train deep learning based object detectors with a number of state-of-the-art models and compare them in terms of speed and accuracy. The fully annotated dataset is released freely available for the research community.  
\end{abstract}

\begin{IEEEkeywords}
Bounding box annotation, object detection, deep learning, indoor dataset 
\end{IEEEkeywords}

\section{Introduction}
%\noindent
Object detection from images is a well-known area of research in machine learning and computer vision. Today, object detection algorithms have matured enough to solve real-world problems. Object detection is a central component in face detection, object counting, visual search, landmark recognition, satellite image analysis, autonomous driving, drone and agriculture production assessment \cite{object_detection_in_agriculture,Object_detection_in_optical_remote_sensing_images, Practical_object_recognition}. Object detection is known to be a challenging task in computer vision as a large number of labeled datasets is needed for learning and generalization performance of the detection model.

The collection of a large scale dataset representative enough is a challenge and the key question in order to train detectors robust to variations in object appearance. Moreover, the annotation of large collections of images is both labor-intensive and error-prone. Traditionally, the annotation problem is solved by brute force, \textit{i.e.,} by crowdsourcing a large group of annotators on a web platform such as the Amazon Mechanical Turk.  Examples of some popular large-scale datasets for object detection with labeled data are ImageNet\cite{Imagenet}, %CIFAR10, CIFAR100\cite{} ,
MS COCO\cite{coco_dataset}, and PASCAL VOC\cite{pascal_voc}. There are also public datasets for specific domains, such as face detection\cite{fddbTech}, character recognition \cite{mnisthandwrittendigit}, landmark recognition and detection \cite{google_landmark}, MCIndoor20000\cite{MCIndoor_dataset} and Freiburg grocery dataset\cite{Freiburg_groceries}.

%Effective and efficient image annotation is one of the major interest among researchers working actively on supervised machine learning and computer vision. To collect a large scale labeled dataset from the specific environment, it is challenging in terms of resource, time and expense. 
As machine learning methods and platforms develop into a more mature state, the focus is turning towards applications. In a small scale application project, the collection of data can not be scaled up by using thousands of annotators. Instead, there is a need for agile and rapid annotation procedures that enable the deployment of machine learning and object detection in small projects, as well. Like most machine learning topics, the most time-consuming task of object detection is also the annotation of each object area in the image dataset. For example, annotating the bounding boxes of a single image from the 14 million sample Imagenet\cite{Imagenet} dataset takes 42 seconds per bounding-box by crowd-sourcing using the Mechanical Turk annotation tools\cite{crowdsourcing}.

There have been many studies focusing on how to speed up the image dataset annotation such as box verification series\cite{box_verification}, % point annotations\cite{click_supervision}, 
eye-tracking \cite{eye_tracking} and learning intelligent dialogs\cite{learning_intelligent_dialogs}. Moreover, semantic annotation of objects in image datasets has been widely discussed in recent years: Polygon RNN++\cite{Polygon_rnn} and Extreme clicking\cite{Extreme_clicking} are recent proposals for efficient object segmentation on image datasets. However, bounding box annotation is by far the most common task in practical and industrial applications, and despite the aforementioned sophisticated approaches, their use is still not very common as easy-to-adopt robust used-friendly tools are missing. 

\begin{figure}[!t]
\centerline{\includegraphics[scale=0.38]{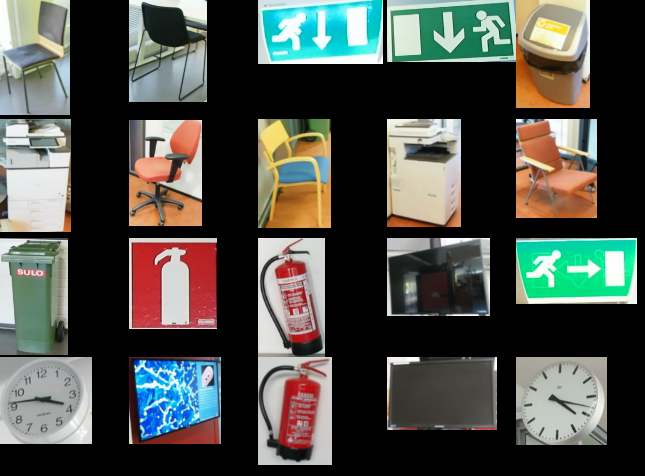}}
\caption{Examples of object instances from TUT indoor dataset.}
\label{example}
\end{figure}

In this work, we study a simple and practical heuristic to annotate the object bounding boxes of an image dataset. Our two-stage approach splits the data into two folds; the first fold is manually annotated from scratch, after which an object detector is trained for generating proposal annotations for the second fold. Thus, the total workload consists of the first stage annotations plus the corrections required to the second fold semi-automatically annotated proposals, and a natural question is to find the optimal split between the folds in order to minimize manual work.

The proposed method significantly reduces the workload to create big enough environment specific object detection dataset. In addition to the technique to fasten the bounding box annotation, we present fully annotated multiclass object detection dataset from indoor scenes, for which we have applied  the proposed annotation procedure. Compared to other indoor datasets, our collection has more class categories, different backgrounds, lighting conditions, occlusion and high intra-class differences. Examples of objects in the dataset are shown in Figure~\ref{example}.

%In this paper, the technique to fasten the bounding box annotation on images is presented together with the fully annotated multiclass object detection dataset from university indoor environment. We named our dataset as TUT indoor dataset. The proposed method is: (1) efficient and economic to collect large-scale labeled dataset %to train the deep learning object detection models,
%, (2) the manual workload can be minimized significantly and (3) provided fully annotated multiclass dataset is handy to do a fast experiment on object detection algorithms.

The remainder of the paper is structured as follow: Section II will describe the dataset. Section III will describe the methods used to fasten the bounding box annotation on image dataset. Section IV will show the experiments and results obtained with them and comments on findings, and Section~V will present the conclusion and future work.

\begin{figure}[tb]
\centerline{\includegraphics[scale=0.70]{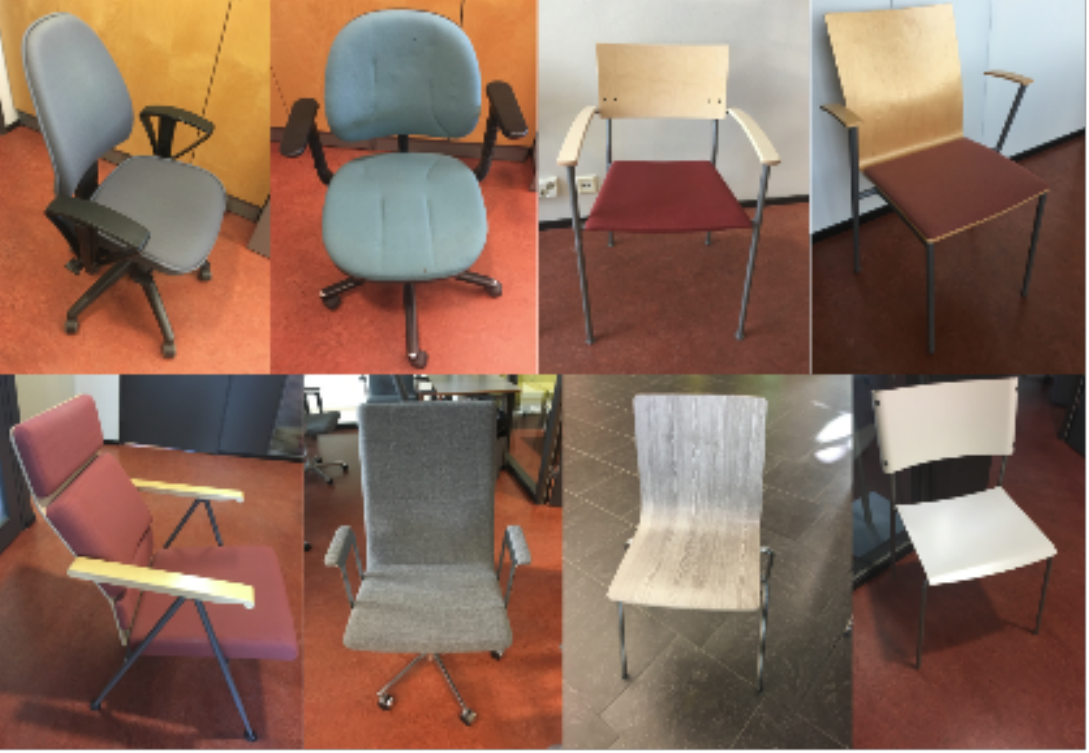}}
\caption{Example of object instances from single class label (\textit{Chair}). Intra-class difference between the object instances are high in \textit{TUT indoor dataset}.}
\label{chairs}
\end{figure}

\section{Dataset}
%\noindent
We have collected an indoor scene dataset to experiment the feasibility of our method. The dataset is created from sequences of videos that were recorded from different indoor premises of Tampere University of Technology (TUT).  The \textit{TUT indoor dataset} is a fully-labeled image dataset to facilitate the board use of image recognition and object detection in indoor scenarios. In addition to the labeled images, we provide the sequence of recorded videos from multiple places inside the university. Our recorded HD videos together with the fully annotated dataset are freely available here \footnote{\url{https://sites.google.com/view/bishwoadhikari/dataset}}.
%[link to data repository]\textbf{TODO Bishwo: create a dummy web page "under construction" to google sites or something. We can change this later.}.
% 

The TUT indoor dataset consists of 2213 image frames containing seven classes. In contrast to existing indoor datasets, our dataset includes a variety of background, lighting conditions, occlusion and high inter-class differences. As shown in Figure \ref{chairs}, the interclass variation is high inside each class category. The variability in terms of background, illumination, blurring, occlusion, pose and scale are main features of TUT indoor dataset.

\begin{figure}[tb]
\centerline{\includegraphics[scale=0.60]{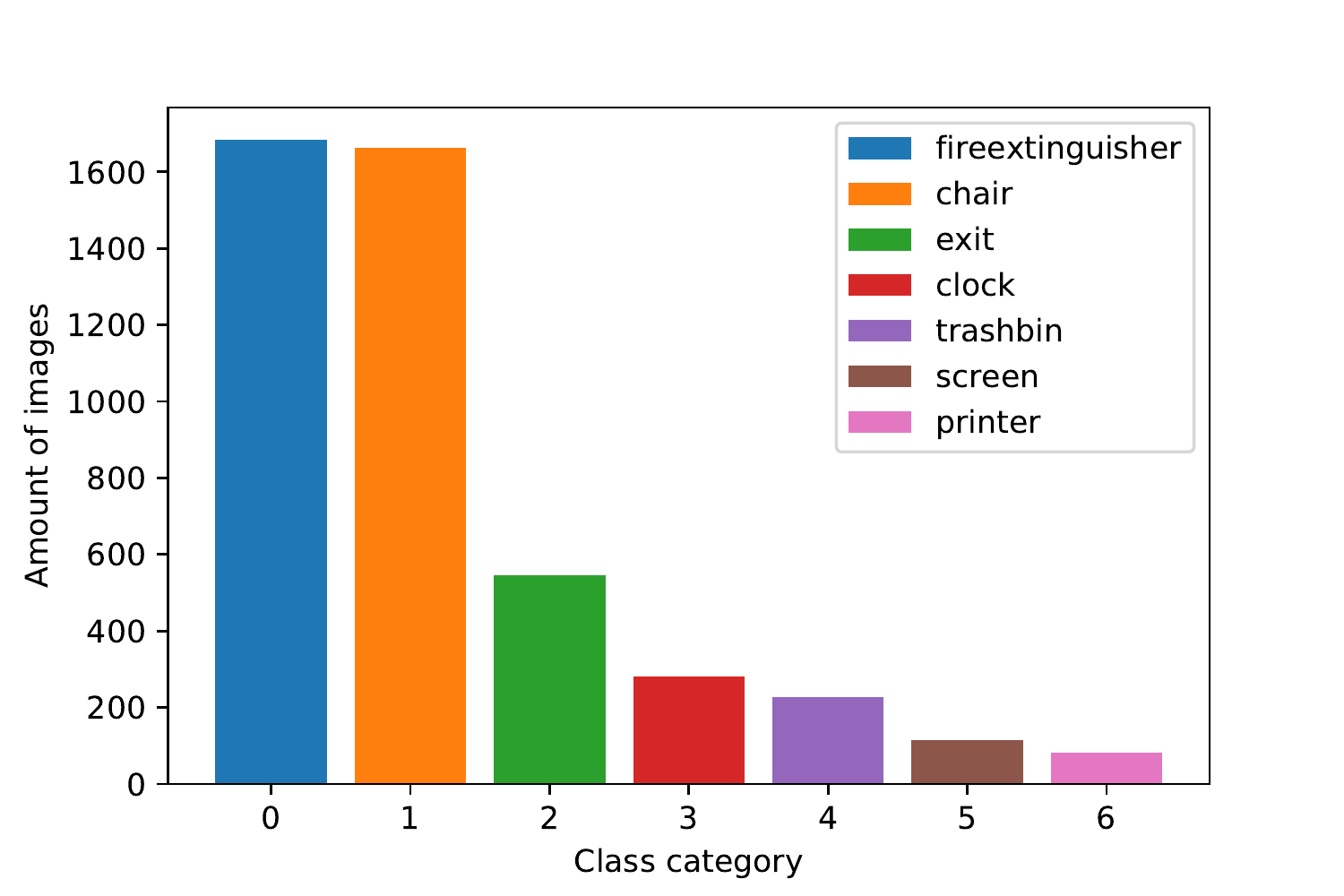}}
\caption{Distribution of the class categories in the \textit{TUT indoor dataset}. There are altogether 4595 instances from 7 categories.}
\label{distribution}
\end{figure}

The size of each frame extracted from the HD video sequences is  $1280 \times 720$ captured using a high-quality camera with optical image stabilization. There are altogether 2213 frames having 4595 object instances from 7 class categories. Each frame consists of several object instances from several class categories. The distribution of object categories is shown in Figure~\ref{distribution}. The highest number of object instances are from \textit{Fire extinguisher} class followed by \textit{Chair} class having more than 1600 instances. While the least populated class category is \textit{Printer} containing less than 100 object instances. The maximum number of object instances from a single class is 1684 and the minimum is 81 instances.

The number of image frames in this dataset can be increased by a factor of 3-4 by using common image data augmentation techniques such as adding noise, blurring, flipping and rotation. As the quality of object detection model and proposal bounding box annotation are directly related to the trained model and images to be annotated, one can improve the performance by improving the quality of bounding box annotation at first place  and increasing the quantity of labeled dataset to train the object detection models. 

\begin{figure*}[tb]
\centerline{\includegraphics[scale=0.60]{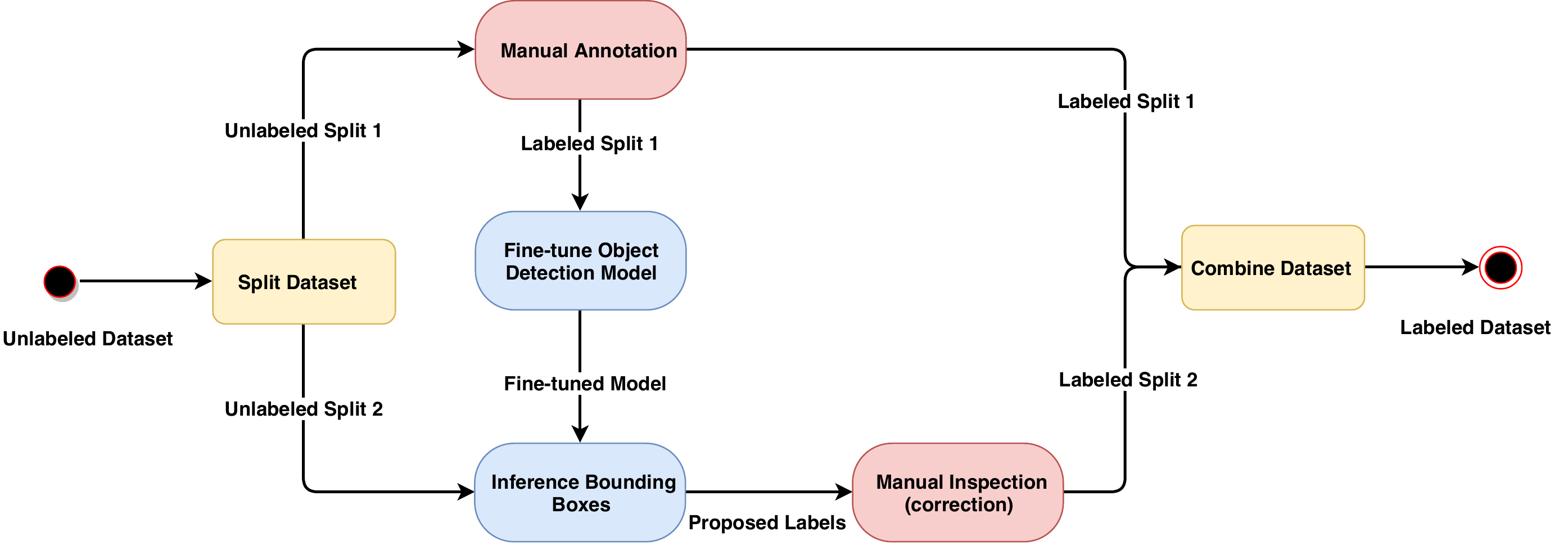}}
\caption{The workflow of our semi-automatic bounding box image annotation method. The dataset is split into two parts. The first part is manually annotated and used to train a model, which is then used to predict labels on the rest of the dataset. After manually correcting the predicted labels, the final fully annotated dataset is combined from the manually annotated and corrected subsets.}
\label{process diagram}
\end{figure*}

\section{Proposed Method for Semi-Automatic Annotation}
%\noindent

In this section, we describe our method for the semi-automatic annotation workflow. The motivation behind the workflow is to minimize the total amount of manual work. The basic idea is to train a model on a small subset of the data, and use that model to predict annotations in the larger set, thus allowing the human annotator to only correct the incorrect predictions in most of the images. More specifically, the workflow consists of six steps as illustrated in Figure~\ref{process diagram}. 

\subsection{Annotation Procedure}

First, the unlabeled dataset is split into two parts. The first (usually smaller) part is first manually annotated by a human from scratch. Then the first part is used to train an object detector, which is then used to predict annotations for the remaining part of the data. The predicted data is manually corrected by a human, after which both parts of the split data are combined to form a fully labeled dataset. Note that two actions out of the outlined six are performed manually by a human, while the other actions can be fully automated. Although the approach is probably invented several times and in practical use already, a systematic evaluation of how this should be done has been dismissed until now. More specifically, we are interested in the proportions the two folds should have in order to minimize manual work and annotation time.

\textbf{\em Splitting the dataset---}The first step in the workflow is to split the dataset to the train and test subsets. The split is performed within the individual video sequence of the TUT indoor dataset, so train and test subsets contain a similar ratio of images from each video sequence. The splitting within the sequences was done in the order of video progression, not randomly, so for instance the first image in every sequence was included in the train set of all the different folds. We experimented with train set folds of 1 \% to 10 \% with 1 \% increases, and 15 \% to 95 \% with 5 \% increases. The corresponding test set for each train set was always the remaining percentage, so a total of 100 \% of the data was used in the experiment. Note that only a subset of the results are shown in the tables.

\textbf{\em Manual annotation of the train set---}%
The second step is to fully annotate the unlabeled fold 1 dataset. This is done by a human. We used a basic bounding box annotation method with no extra speed up procedures.

\textbf{\em Training the detector---}%
The third step is to train the detector. Although any detector can be used, we focus on the recent deep learning based object detection models, which we fine-tune using the manually annotated dataset. More specifically, we choose the Faster RCNN model using ResNet-101 network trained on the MS COCO dataset as our starting point. The pretrained models for object detection can be found on the TensorFlow Object Detection model zoo \footnote{\url {https://github.com/tensorflow/models/blob/master/research/object_detection/g3doc/detection_model_zoo.md}}.

\textbf{\em Predicting annotations for the rest of the data---}%
After fine-tuning the object detection model, it is used to predict the  bounding boxes for the rest of the unlabeled dataset, the unlabeled fold 2. 

\textbf{\em Manually correcting the inferred proposals---}%
The inferred predictions are manually corrected by a human. The human needs to go through all the proposed labels, see if they make sense, and correct if necessary. Wrongly drawn boxes are removed, wrongly labeled classes are corrected and new boxes are drawn, if needed. We used the same annotation tool as when fully annotating.

\textbf{\em Combining the final labeled dataset---}%
The fully annotated train fold and the fully corrected test fold are finally combined as the labeled dataset. 

\subsection{Estimating the Workload}
%\noindent

The total amount of manual work consists of three kinds of manual operations: 
\begin{enumerate}
\item Annotation of bounding boxes in the first fold,
\item Removal of false detection (false positives) in the second fold,
\item Addition of missed detections (false negatives) in the second fold.
\end{enumerate}
We choose to define the false negatives and false positives by the amount of overlap between the true and predicted bounding box: We assume the user would correct the annotation if the \textit{intersection-over-union} (IoU) overlap between the true object location and the predicted bounding box is less than 50 \%. 

Additionally, in the case of partial overlap less than 50 \%, we model the user operation as removal of the incorrect box and addition of the box at the correct location. Alternatively, one might consider an additional user action: moving and/or resizing the box. However, this requires often more work (and mental attention) than simple removal and addition. Moreover, many annotation tools have not even implemented the box adjustment operation.

The workload required for the second fold corrections can be estimated from the precision and recall values of the first stage object detector. Precision is defined as 
\[
\text{precision} = \frac{\text{\# of correct detections}}{\text{\# of all detections}},
\]
and recall as
\[
\text{recall} = \frac{\text{\# of correct detections}}{\text{\# of all objects}}.
\]
%the sum of the correctly detected objects divided by the total population of the object that is detected by the detector. Recall is the sum of the correctly detected objects divided by the sum of the actual objects. The precision and recall values are based on the intersection over union (IoU) measure. An IoU is calculated as the ratio of the area of overlap and area of the union between the true box and the proposed box. 
Using these two metrics, we can estimate what would be the workload for the tasks 1-3 above, respectively.

\textbf{\em First stage annotations---}In the first stage, the user simply marks each object in each first fold image. Therefore, the number of manual operations is simply
\[
\text{\# initial annotations} = \text{\# of true objects in fold 1}.
\]

\textbf{\em Second stage additions---}The recall value summarizes the proportion of true objects in the second stage fold that are correctly found by the detector. The complement of recall is then the proportion of objects \textit{not} found by the detector. Thus, the number of additions the user would have to perform is given as
\[
\text{\# additions} = (\text{\# of true objects})\times  (1 - \text{recall}).
\]

\textbf{\em Second stage removals---}On the other hand, the user has to remove all false detections, \textit{i.e.,} detections at places with no true object. The precision value describes the proportion of detections that are in fact true objects. The complement of precision is then the proportion of detections that do \textit{not} correspond to a true object, and needs to be manually removed. Thus, the number of removals the user would have to perform is given as
\[
\text{\# removals} = (\text{\# of all detections})\times  (1 - \text{precision}).
\]

The total work is the sum of these three steps:
\[
\text{operations} = \text{initial annotations} + \text{additions} + \text{removals}.
\]
However, we will see that the amount of time required for the different tasks varies a lot. In particular, the box removal is very fast, while the annotations from scratch take more time. Therefore, we also define the total working time required:
\[
\text{time} = t_1 \left(\text{initial annotations}\right) + t_2 \left(\text{additions} + \text{removals}\right),
\]
where $t_1$ denotes the time required for a single 1st stage annotation, and $t_2$ is the average time for a single 2nd stage correction.

%Result Table (NEW)
\begin{table}[tb]
\renewcommand{\arraystretch}{1.4}
\caption{Performance evaluation of fine-tuned tensorflow object detection api models and retinanet models using tut indoor dataset.}
\label{performance}
\centering
\begin{tabular}{l|c|c}
\hline
\bfseries Model & \bfseries mAP \% & \bfseries Speed (FPS)\\
\hline\hline
TensorFlow   SSD MobileNet 			& \bfseries 97.73	& \bfseries 27.41 \\
TensorFlow   Faster RCNN Resnet50 	& 96.54 	& 11.80\\
TensorFlow   Faster RCNN Resnet101 	& 95.06  	& 9.69 \\

\hline
RetinaNet,  backbone = Resnet50 	& 96.60  	& 13.00 \\
RetinaNet,  backbone = Resnet101 	& 95.93  	& 9.64 \\
RetinaNet,  backbone = VGG16  		& 95.67 	& 9.79 \\
RetinaNet,  backbone = VGG19  		& 94.57  	& 8.66 \\ 
\hline
\end{tabular}
\label{tab:1}
\end{table}

\section{Experimental Results}
%\noindent
Our interest is to estimate the workload needed to create a fully annotated environment specific multiclass object detection dataset. However, as we are also introducing a new dataset not previously studied, we will first assess its difficulty by training state-of-the-art object detectors and evaluate their accuracy. Understanding the accuracies of different models will also help to evaluate the impact of choosing a poor object detector for the second stage proposal generation. 
After that, we will study the annotation workload that would be required to annotate the data using different strategies.\footnote{To clarify: the dataset is fully annotated, but we will compare different strategies how this could have been done faster.}

%At first the workload is measured in number of bounding boxes that are needed to be drawn and then compute the total time in seconds needed for whole dataset annotation. 

%We focus on evaluating our method when the number of labeled images is much smaller than that of unlabeled ones. We randomly split the dataset into the 10\% subset for training and the 90\% subset for testing. We repeat the same experiment on different splits of training and testing and calculated the workload on those splits.

\textbf{\em Accuracy on the TUT indoor dataset---}For this study, we use three recent object detection pipelines. First one is the Regions-CNN (R-CNN) framework proposed by Ren \textit{et al.} in 2015\cite{ren2015faster}, which uses a two-stage structure: first stage network creates object proposals, which the second stage network then classifies into different categories. The second approach is the Single Shot Detector (SSD) framework proposed in 2016 by Liu \textit{et al.} \cite{liu2016ssd}. Instead of the two-stage structure, SSD uses only single feedforward network to predict object locations (regression) and categories (classification) is a single network. Finally, we consider the \textit{RetineNet} structure proposed by Lin \textit{et al.} in 2017 \cite{lin2017focal}. The RetinaNet extends the SSD approach by defining a novel loss function in order to focus the attention to the most difficult cases instead of the easy ones.

%Result Table
\begin{table}[tb]
\renewcommand{\arraystretch}{1.2}
\caption{The total manual annotation workload at different dataset splitting ratios. A bigger training set results in better inference performance, as indicated by the mAP. For our dataset, the best workload is found when the training set size is between about 4\% - 10\%.}
\label{workload}
\centering
\begin{tabular}{l|c|c}
\hline
\bfseries Split & \bfseries Workload & \bfseries COCO mAP \% @ IoU 0.5\\
\hline\hline
Train 1 \%, Test 99 \% & 1642 & 56.88 \\
Train 2 \%, Test 98 \% & 1118 & 67.45 \\
Train 3 \%, Test 97 \% & 970 & 75.99 \\
Train 4 \%, Test 96 \% & 915 & 77.75 \\
Train 5 \%, Test 95 \% & 943 & 80.18 \\
Train 6 \%, Test 94 \% & 863 & 81.63 \\
Train 7 \%, Test 93 \% & 938 & 82.56 \\
Train 8 \%, Test 92 \% & 979 & 85.46 \\
Train 9 \%, Test 91 \% & 931 & 79.96 \\
Train 10 \%, Test 90 \% & 963 & 82.72 \\
\hline
Train 20 \%, Test 80 \% & 1265 & 88.80 \\
Train 40 \%, Test 60 \% & 2016 & 95.42 \\
Train 60 \%, Test 40 \% & 2878 & 96.76 \\
Train 80 \%, Test 20 \% & 3704 & 96.65 \\ 
\hline
\end{tabular}
\label{tab:2}
\end{table}

For all cases, we start training from MS COCO trained network and fine-tune using our own data. For the R-CNN and SSD structures we use the TensorFlow object detection framework\cite{tensorflow_object_detection_api} and for RetinaNet we use the Keras-RetinaNet library\footnote{\url{https://github.com/fizyr/keras-retinanet}}.

The accuracies of the  experimented models are shown in Table \ref{tab:1}. In this experiment, 80 \% of the total dataset is used for training the model and the rest is used for evaluating the performance of the detector. %It is obvious that the SSD MobileNet model is faster than rest of the experimented models.
The mean average precision (mAP) calculation in our experiment follows the MS COCO \cite{coco_dataset} evaluation procedure, \textit{i.e.,} we use a fixed set of 101 detection thresholds $t=0,0.01,\ldots , 1.0$ with no interpolation and average the precision metrics over all thresholds. We consider matches only for the IoU value of 0.5 and over, and not averaged over several IoUs as in the typical COCO evaluation metric.

According to the results of Table \ref{tab:1}, the simplest SSD network seems to have both the highest accuracy and highest computational speed. This is somewhat surprising as two-stage detectors (R-CNN) tend to have higher accuracy than the single-stage detectors\cite{retinanet}. On the other hand, this is good news since there is no need to consider the tradeoff between the speed and accuracy. 

% %Result Table (OLD)
% \begin{table}[H]
% \renewcommand{\arraystretch}{1.4}
% \caption{Performance evaluation of fine-tuned tensorflow object detection api models and retinanet models using tut indoor dataset.}
% \label{performance}
% \centering
% \begin{tabular}{l|c|c}
% \hline
% \bfseries Model & \bfseries mAP \% & \bfseries Speed (FPS)\\
% \hline\hline
% TensorFlow   SSD MobileNet & 83.04  & \bfseries24.24 \\
% TensorFlow   Faster RCNN Resnet101 & 95.42  &9.12 \\
% TensorFlow   Faster RCNN Resnet50 & 95.05 &9.27\\
% \hline
% RetinaNet   backbone = Resnet50 & \bfseries 97.36  &12.39 \\
% RetinaNet   backbone = Resnet101 & \bfseries 97.36  &8.55\\
% RetinaNet   backbone = VGG16  & 96.79 &8.96 \\
% RetinaNet   backbone = VGG19  & 96.92  &8.55 \\ 
% \hline
% \end{tabular}
% \end{table}

\textbf{\em Workload minimization---}%
The workload calculation based on different train-test splits is summarized in Table~\ref{tab:2}. In this case, we use the Faster RCNN with ResNet101 for generating the second stage proposals. We decide to use this network instead of the fastest and more accurate SSD network because the SSD performance might be slightly anomalous and specific to our dataset only. Moreover, the performance differences are minor, so essentially any of structure would produce more or less similar results.
In each case, we train altogether 50000 epochs.

\begin{figure*}[!t]
	\subfloat{\fbox{\includegraphics[scale=0.57]{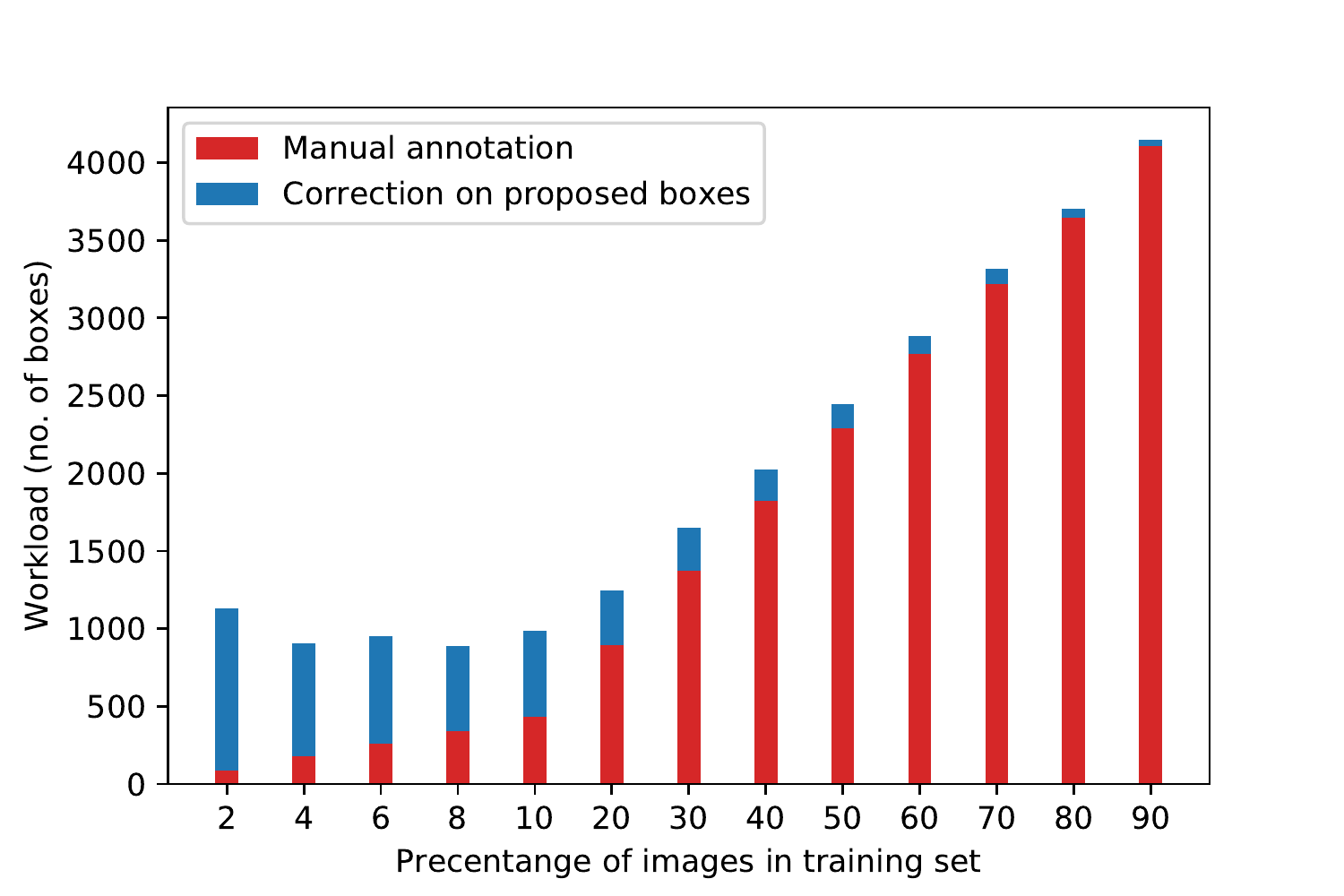}}}
    \subfloat{\fbox{\includegraphics[scale=0.57]{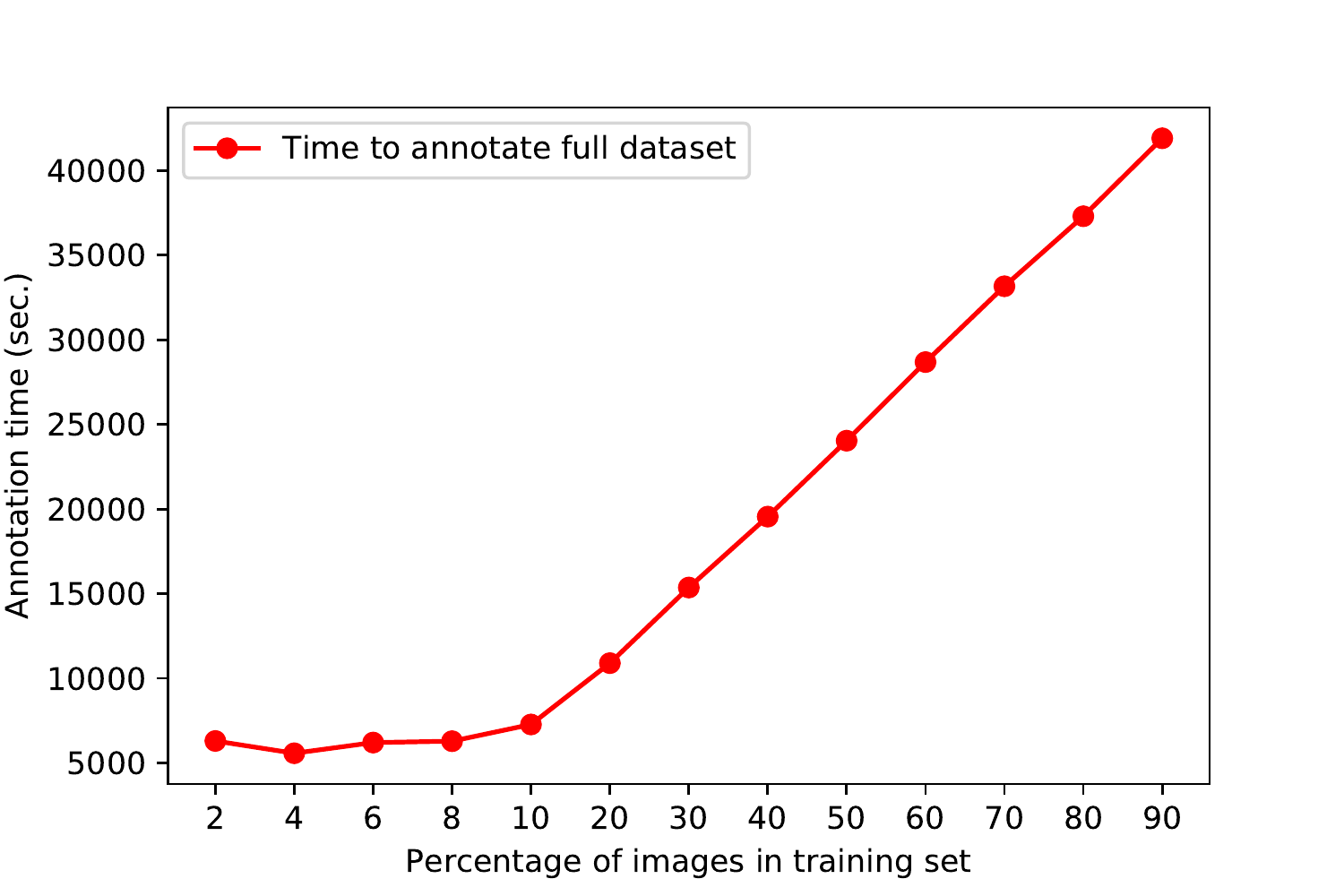}}}
    \caption{Amount of workload needed in different train-test split (left). The proportion of annotation time needed to annotate full dataset using different portion of manually annotated data to fine-tuned the object detection model (right).}
    \label{results_fig}
\end{figure*}

The total manual workload to create a fully annotated dataset is the sum of manual workload for annotating the first dataset split and the workload needed for the correction. %The mAP illustrates the inference quality of the model on the corresponding test set split, when it is trained with the corresponding training set split. 
To minimize the workload, the initial annotation workload should be as low as possible, and the quality of the model as high as possible. The optimal minimized workload is found when these two conflicting requirements are both satisfied well enough.

To make the workload figures more intuitive, we also consider the time spent in the manual stages of the annotation process, which are based on the workload figures. From our timing experiment, we found that manual annotation takes about 10.15 seconds per bounding box and correction (addition and removal) takes about 5.20 seconds per bounding box. (In the timing experiment we used 80\% of total dataset on training the model and remaining 20\% to predict the bounding boxes). We are using these approximate times to estimate the time needed to draw all together 4595 bounding boxes in all image frames in TUT indoor dataset.

The calculation of the total time for the annotation is: 
\begin{equation*}
\text{Total time for annotation} = W_M \times 10.15 + W_C \times 5.20 
\end{equation*} 
Where, $W_M$ is number of boxes required to annotate in the first fold of the dataset and $W_C$ is workload for the correction of proposed bounding boxes in fold 2. 

Figure \ref{results_fig} summarizes the results of work required to annotate the whole dataset. On the left, we show the number of bounding boxes to annotate in the two stages, as a function of the proportion of first/second stage folds. One can see that with very small first stage fold, the first stage becomes very easy to annotate, but the second stage proposals become unreliable and more work is needed on the second stage. On the other hand, if excessive amounts of training data is allocated to the first stage, then the total workload is dominated by the first stage, while the seconds stage requires virtually no corrections. The sweet spot is at a relatively low percentage: The best strategy is to annotate only 4--8\% from scratch and use the trained model for the rest.

In the right panel of Figure \ref{results_fig} we show the time used for the total annotation in different split proportions, \textit{i.e.,} we multiply the number of boxes with their estimated execution times. This has the effect of moving the sweet spot even further to the left, with the minimum at 4\%. 

In all splits, our method decreases the manual workload to annotate the full dataset significantly. Based on our measured annotation time, the total time needed to annotate the full TUT indoor dataset manually from scratch would be $4595 \times 10.15 = 46639.25$ seconds $\approx$ 13 hours.  
%As shown in left graph of Figure 5, the workload needed to annotate the full dataset varies on the portion of data used to fine-tune the object detection model. However, in all cases it is less than the total estimated time required for full manual work by human. In our case, we found that the minimum workload is achieved while training with 4 - 10\% of whole dataset and inference on the remaining portion.
Using the proposed two-stage approach the required time decreases to less than two hours of manual labor.%, as illustrated in the right side graph of Figure 5. 
Also, the total work time starts to increase quite linearly with train set splits higher than 10\%. Note that the time estimations include only the manual labor done by humans, and do not include the automated computing time between the manual phases of the task.

\section{Conclusions}
%\noindent
In this paper, we introduced the efficient way to annotate the object detection dataset using fine-tuned object detection model. %The main point is to inference bounding boxes and object class on images. 
We introduced a fully labeled TUT indoor dataset for object detection. Our dataset have 2213 image frames extracted from 6 different sequences of recorded videos containing 7 classes of common indoor objects. We tested the relevance of the dataset with current state-of-the-art object detection frameworks. The real-world indoor dataset is valuable resource for indoor object detection and handy for the fast experiment on object detection algorithms.

% More specifically, the automatic annotation technique for the bounding box annotation is of major interest. As data annotation is one of the challenging factor for deep learning, automating the image annotation helps to reduce the time and cost constraints to implement deep neural network model in computer vision problems. 

The main contribution of the paper is a two-stage procedure for rapid annotation of bounding boxes, together with a systematic assessment of the time savings introduced by the approach.
It was found that the proposed two-stage method reduces manual work by almost 90 \% and is cost efficient for the implementation of supervised object annotation campaigns. 

We experimented that manual annotation takes about 10.15 seconds per bounding box while the correction takes about 5.20 seconds per box. Note that this calculation might differ in different scenarios such as different object types, changes in number of classes and annotation tools.
It is found that the workload to annotate TUT indoor dataset  is minimum when using 4--8\% of the total dataset to fine-tuned the object detection model. %This will be 9 times faster to labeled all images in TUT indoor dataset than fully labeling by human. 

% The total workload required to created the fully labeled object detection dataset is reduced to .. \% of manual workload needed to labeled all images in the dataset. Notice that the amount of workload might be differ in different scenarios. 

For future work, we plan to  experiment more with other datasets to see how the procedure generalizes to larger number of classes, for instance. Secondly, instead of two stages, we plan to generalize the approach to more stages, incrementally improving the proposal model accuracy at each stage. 

\section*{Acknowledgment}
The authors would also like to thank CSC - The IT Center for Science for the use of their computational resources. The work was partially funded by the Academy of Finland
project \textit{309903 CoefNet} and Business Finland project \textit{408/31/2018 MIDAS}.

% \bibliography{references.bib}
% \bibliographystyle{IEEEtran}

% \begin{figure}[!t]
% \centerline{\includegraphics[scale=0.40]{examples.png}}
% \caption{Examples of object instances from TUT indoor dataset.}
% \label{example}
% \end{figure}

% Generated by IEEEtran.bst, version: 1.14 (2015/08/26)

\end{document}